\documentclass{article}

\usepackage{PRIMEarxiv}

\usepackage[utf8]{inputenc} 
\usepackage[T1]{fontenc}    
\usepackage{hyperref}       
\usepackage{url}            
\usepackage{booktabs}       
\usepackage{amsfonts}       
\usepackage{nicefrac}       
\usepackage{microtype}      
\usepackage{lipsum}
\usepackage{fancyhdr}       
\usepackage{graphicx}       
\usepackage{caption,subcaption}
\usepackage{float}
\newcommand{\etal}{\textit{et al.}}
\usepackage{todonotes}

\title{Animating Still Images
}

\author{
  Kushagr Batra \\
  Bal Bharati Public School \\
  \texttt{kushagr.batra@gmail.com} \\
   \And
  Mridul Kavidayal \\
  University of California San Diego \\
  \texttt{mridul.kavidayal@gmail.com} \\
}

\begin{document}
\maketitle

\begin{abstract}
We present a method for imparting motion to a still 2D image. Our method, uses deep learning to segment a section of the image denoted as \textit{subject}, then uses in-painting to complete the background, and finally adds animation to the subject by embedding the image in a triangle mesh, while preserving the rest of the image. 
\end{abstract}


\section{Introduction}
Digital media has become ubiquitous with modern life. Digital images surrounds us in all works of like. With the advent of mobile phone cameras, the number of digital images created has grown immeasurably. Most of the platforms provide different ways of enhancing the creative expression of an individual by providing various effects. 

Machine learning and AI have pushed the boundaries of edibility of 2D images. One may choose from a plethora of available filters, starting from the elementary, such as adding shadows, changing hues, to the very complex where even the style of the photos can be altered.

In this paper, we propose a motion effect by exploiting some of the advanced techniques in deep learning and computer vision. We show how a 2D digital image can be transformed into a short movie, capturing the dynamics of the scene and creativity of an individual.
\section{Background}
\label{sec:headings}
We have solved the problem of imparting motion to 2D images via decomposition into a sequence of vision problems. In this section, we provide a gentle introduction to these problems as well as mention the available state-of-the-art solutions.  
\subsection{Image Segmentation}
Image segmentation divides an image into many distinct regions depending on certain characteristics. The division could either be based purely on visual cues (like color, brightness, gradient, etc.) or semantic cues (objects). The process involves labelling pixels based on certain requirements to produce segmented regions. This section describes few of the image segmentation techniques. 
\subsubsection{Edge Detection}
These techniques generally involve strategies to detect edges and form regions/segments using some heuristics. The assumption is that there is always an edge between two adjacent regions with different gray-scale values. \textit{Sobel filter} is a commonly used filter to detect edges. It detects sudden changes in pixel intensity which are characterized as edge. Canny edge detector~\cite{canny} operator is yet another popular filter for identifying edges. 

\subsubsection{Clustering}
Clustering algorithms such as k-means can be used to generate pixel clusters based on visual cues. The algorithm works be repeated performing the following steps:
\begin{itemize}
    \item Select  k initial center points. These can be user guided or random.
    \item Find the center point which is most representative of them. This could be either done by finding their distance or in our case for image segmentation, the rgb pixel values could be used.
    \item Once, all image pixels are assigned a label from the initial k classes, new centers are then chosen for these centers (which could be the average rgb values here) while subsequently merging clusters which are close (the metric for closeness would again be rgb values).
\end{itemize}
These steps are repeated until no clusters are merged.
\subsubsection{Feature learning}
These techniques involve \textit{generating} (scale-invariant feature transform, or SIFT) ) or \textit{learning} (Deep Neural Networks) features and using them for segmentation, recognition, etc. \textbf{SIFT}~\cite{Burger2016} helps locate the local features in an image, commonly known as the keypoints of the image. These keypoints are scale and rotation invariant that can be used for various computer vision applications, like image matching, object detection, scene detection, etc. The major advantage of SIFT features, over edge features is that they are not affected by the size or orientation of the image. 

More recently deep neural networks have worked extremely well in learning and extracting useful features from images. The basic idea is to extract features by convolving the image with weights. Earlier methods like \emph{Sobel} used hard-coded weights for identifying features (vertical or horizontal edges in this case). With neural networks, these weights are learned during the training phase according to the input data. So if the data consists of images of dogs and cats, weights would be learned so that some neuron (perceptron) fire up only when features of a cat are detected and some fire up only when features of a dog are detected. The loss defined during training enables the network to learn these weights. 

Couple of recent deep neural networks that have worked well for the task of image segmentation are as follows:
\begin{itemize}
    \item \textbf{U-Net}~\cite{UNET2015}, a convolution neural network whose architecture looks like the letter U and hence the name U-Net. Its architecture is made up of two parts, the left part — the contracting path and the right part — the expansive path. 
    \item \textbf{Mask R-CNN}~\cite{mrcnn2017}. In this architecture, objects are classified and localized using a bounding box and semantic segmentation that classifies each pixel into a set of categories. Every region of interest gets a segmentation mask. A class label and a bounding box are produced as the final output. 
\end{itemize}

\subsection{Inpainting}
Image inpainting is a field of research which involves filling in the missing area within a picture with meaningful pixels. The goal of this task is to generate the image looking as realistic as possible. In the past few years, deep learning has achieved great success on the image completion. 

Pathak \etal \cite{pathak} generate the missing part of the image using the \emph{Encoder Decoder} structure trained on the adversarial loss. The input is a picture with a missing region in the center. The encoder produces a latent feature representation of that image. The decoder produces the missing image content. The network is trained to match the ground truth of the missing region. The reconstruction loss is an $l2$ loss between ground truth image and the produced image. The adversarial loss is given so that the generator produces the image that tries to make it hard for the discriminator to distinguish between the generated and real images. the architecture depends on the missing region’s shape, which could be inconvenient in a real-world application. 

Yu \etal \cite{Yu} use a two-stage network and a contextual attention module to improve the quality of the generated images. The architecture is described as coarse to fine Network. They make use of a module in the refinement network, and it calculates the similarity within the image. The motivation behind this module is that when you are reconstructing a missing region, you are going to look at the most similar looking region in other parts of the same image. 

Liu \etal \cite{Liu} produce the missing region of the image using an existing U-net + partial convolution. Partial convolution is a convolution operation that uses only the non-masked region. This work is notable because it only adds the partial convolution and combination of popular losses on the existing U-net architecture to generate remarkable quality images. The method does not constraint on the number of holes in the image or their subsequent location. 

\section{Animation}
Animation can be used to bring 2D characters to life through application of a few well-known principles, such as squash, stretch and follow through. Conceptually, frame-by-frame animation is the simplest method, where animation changes the contents of the image in every frame. It is best suited to complex animation in which an image changes in every frame instead of simply moving across the scene. 

Mesh deformations is a method of animation where a static 2D image to be broken down into customized polygons. These polygons can then be stretched or warped by the manipulation of each individual vertex.

There are various digital techniques used to manipulate graphics for animations such as linear blend or dual quaternion skinning techniques (LBS and DQS, respectively).
Linear blend skinning (LBS) offers a simple solution for capturing this behaviour while enabling to smoothly blend deformations near joints.
Being fast and highly customizable, LBS is routinely used in the animation pipeline.

In addition, there are also various deep learning powered techniques which can directly produce animated images with help of reference videos, such as \cite{DBLP:journals/corr/abs-1812-08861}. \cite{DBLP:journals/corr/abs-1812-08861} introduces a framework for motion-driven image animation to automatically generate videos by combining the appearance information derived from a source image (e.g. depicting the face or the body silhouette of a certain person) with motion patterns extracted from a driving video (e.g. encoding the facial expressions or the body movements of another person). Over the past few years, researchers have developed ap- proaches for automatic synthesis and enhancement of visual data. 

\label{sec:motionTransfer}
\section{Method}
We now describe the steps to achieve motion transfer.
\begin{enumerate}
    \item The segment of the image is extracted for imparting motion. 
    We chose \textit{Reviving Iterative Training with Mask Guidance for Interactive Segmentation}~\cite{ritmg2021}. It is a click-based interactive segmentation technique and generates a mask for the desired object by refining it via user's input clicks. There are positive and negative clicks in click-based interactive segmentation. These clicks are represented by their coordinates in an image. Clicks are represented as an image via a distance transform and fed in the network by concatenating to the input image as a separate channel. See figure~\ref{fig:seg}.
\begin{figure}[H]
  \centering
  \includegraphics[width=\textwidth,keepaspectratio]{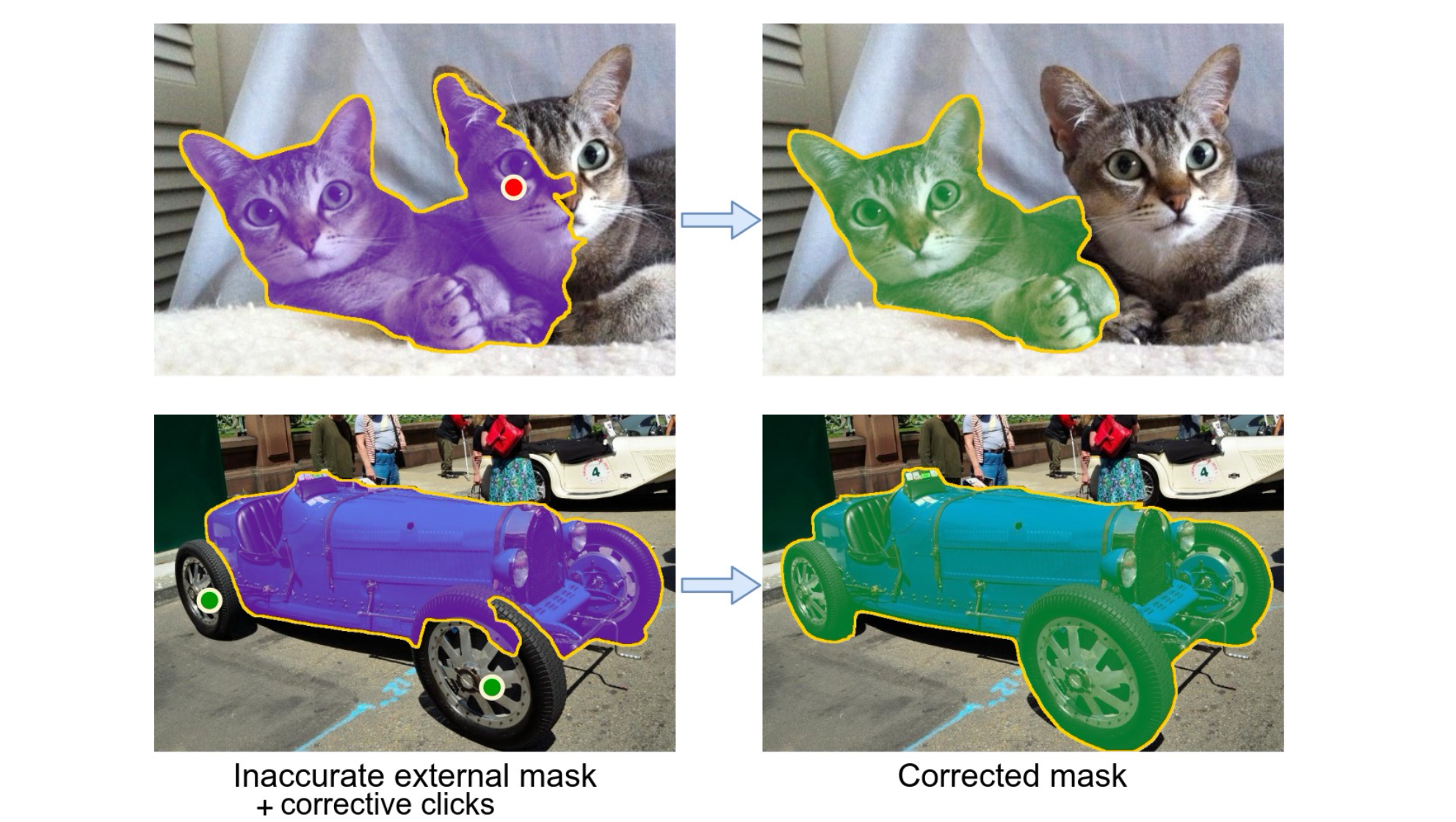}
  \caption{Interactive segmentation: Green markers are used to let the network know about a region of interest. Red markers indicate an incorrect region}
  \label{fig:seg}
\end{figure}
    This will generate a mask for the segmentation. The masked segment of the image is call the \emph{subject} of the image.
    
    \item The image and the mask is used for in-paining to fill the masked segment of the image. This work uses \textit{MAT: Mask-Aware Transformer for Large Hole Image Inpainting} \cite{mat} which employs a transformer-based model for large hole \emph{inpainting}, thereby unifying the merits of transformers and convolutions to efficiently process high resolution images.  
    
\begin{figure}[H]
  \centering
  \includegraphics[width=\textwidth,keepaspectratio]{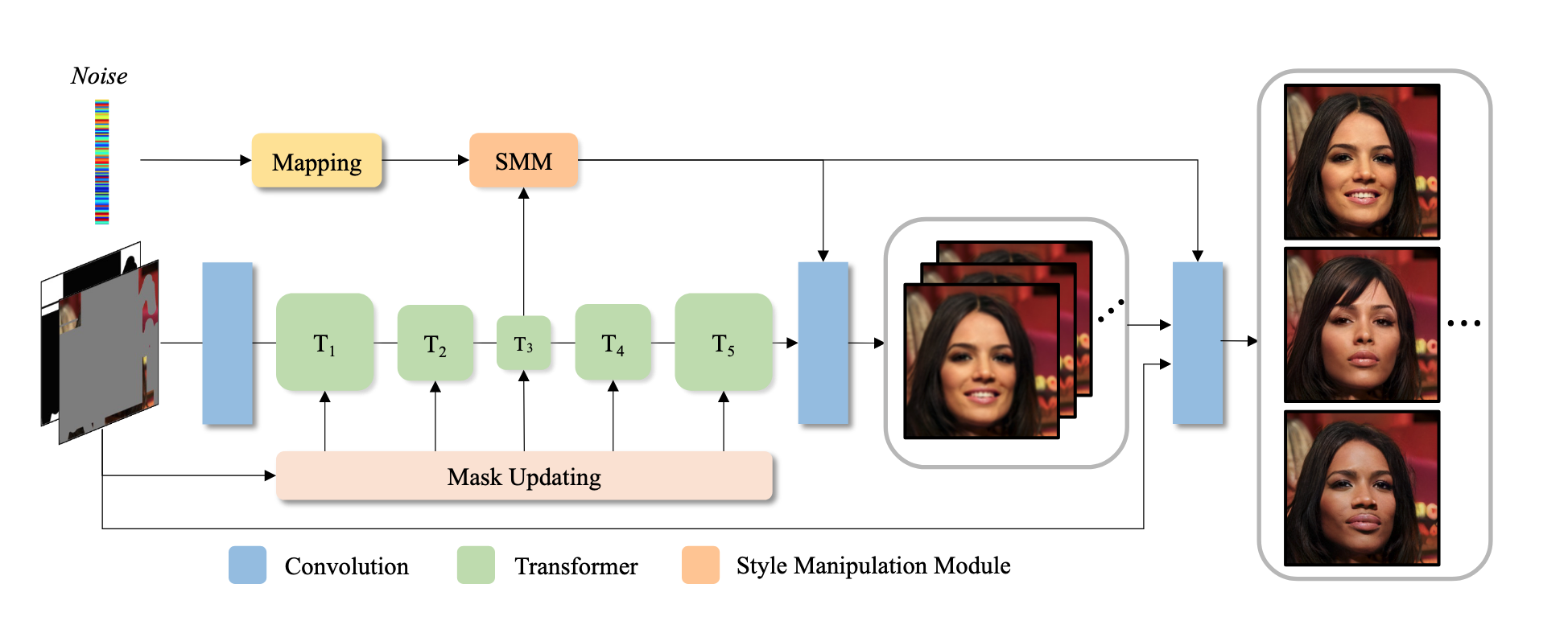}
  \caption{MAT Architecture}
  \label{fig:inp}
\end{figure}
    MAT architecture consists of a convolution head, a transformer body, a convolutions tail and a style manipulation module. The convolution head is used to extract tokens, then the main body with five stages of transformer, models long-range interactions via the multi-head contextual attention (MCA). For the output tokens from the body, a convolution-based reconstruction module is adopted to up-sample the spatial resolution to the input size. A style manipulation module is also used for delivering diverse predictions by modulating the weights of convolutions\cite{hadash2018estimate}. Refer to figure~\ref{fig:inp}. 
    
    \item Using techniques described in previous steps, we can segmented the image, typically into a foreground subject and background. We will be animating/moving the subject while keeping background the same.
    In addition, we have used foreground subject mask to do in-painting on input image. 
    This is useful because when we move the foreground subject, having a hole where the subject used to be looks odd and is undesirable. In-painting completes the subject's region from neighbouring areas and completes the background image.
    
    At this point, we are ready to start animating the subject. To do so, we take our images into the triangle mesh domain. We create two overlapping triangle meshes of suitable density, and use the background image and foreground image as texture on the two triangle meshes, refer Figure \ref{fig:meshTex}. 
    
    \begin{figure}
        \centering 
      \includegraphics[width=0.40\textwidth]{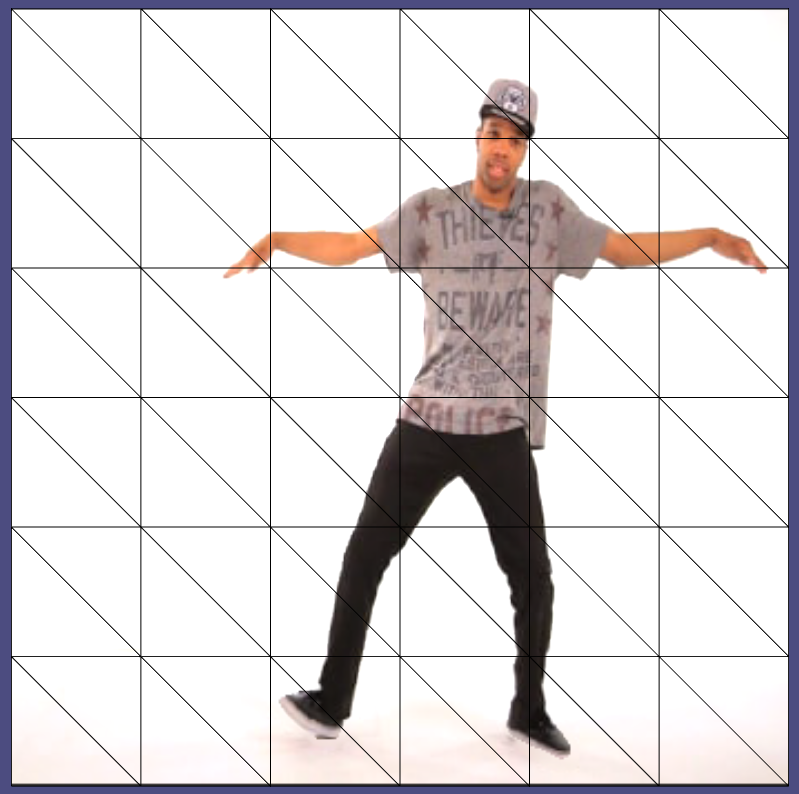}
      \caption{A sample image with texture on triangle mesh.}
      \label{fig:meshTex}
     \end{figure}
    
    Now, for animating our subject, we manipulate the vertices of subject triangle mesh appropriately, while keeping the background mesh as-is. This technique allows for much more flexibility than directly manipulating image pixels to generate new image for every single frame. We allow for the following kind of animations on our subject :
    \begin{itemize}
        \item \textbf{Horizontal Wave:}
        Horizontal waves are implemented as a sine wave travelling along a vertical direction. As the waves travel vertically, it moves the vertices of triangle mesh accordingly in horizontal direction.
        \item \textbf{Vertical Wave:}
        Vertical waves are implemented as a sine wave travelling along a horizontal direction. As the waves travel horizontally, it moves the vertices of triangle mesh accordingly in vertical direction. 
        \item \textbf{Jump:}
        For a natural looking jump, we divide the entire animation into several stages, in between the stages the subject pose is obtained via interpolation of various scale and translation parameters. Various stages are as follows:
        \begin{itemize}
            \item Phase One: This is the rest pose, where the subject image is at rest. 
            \item Phase two: In this stage, subject gets ready for jump. To depict so, we scale subject horizontally to 110\% of original width, and vertically to 90\% of original height.
            \item Phase three: This is the peak of jump, subject is horizontally scaled to 90\% of original width, vertically to 110\% of original height and translated vertically to 50\% of original height of the subject.
            \item Phase four: At this stage, subject is back on the ground, so there is no vertical displacement. However, we animate a second smaller bounce for natural effect, and in preparation of that, subject is scaled to 105\% of original width and 95\% of original height.
            \item Phase five: Small second bounce. Subject's height and width are as original, but it is displaced vertically by 2\% of original height.
            \item Phase six: This is same as rest pose, there is no scaling or displacement of subject.
        \end{itemize}
\end{itemize}

In each of above animations, there are various parameters that can be tweaked to get different animations. For example, number of sine waves in any of the wave animations, it's amplitude etc. Similarly we can tweak the scaling and displacement at various stages of jump animation as required.
\end{enumerate}



\subsection{Results}
We present, some of the results of segmentation, in-painting and finally motion Transfer. Figures below shows few of our input examples.

Full summary of the results, along with generated animation videos is available at \href{https://kushagr2402.github.io/motiontransfer}{https://kushagr2402.github.io/motiontransfer}. 

And the source code is available at \href{https://github.com/kushagr2402/motiontransfer}{https://github.com/kushagr2402/motiontransfer}

\begin{figure}[H]
    \centering 
\begin{subfigure}{0.3\textwidth}
  \includegraphics[width=\linewidth]{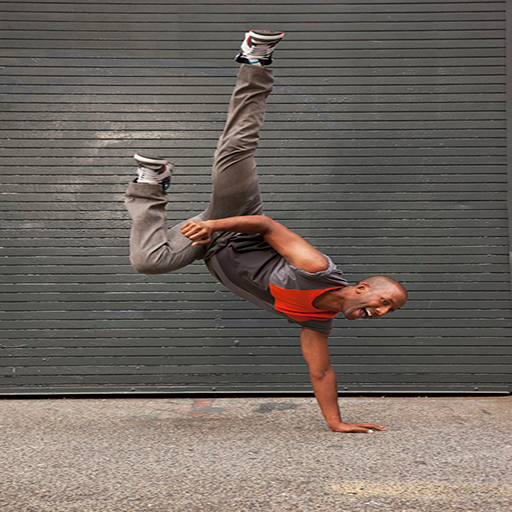}
  \caption{Input Image}
  \label{fig:res11}
\end{subfigure}\hfil 
\begin{subfigure}{0.3\textwidth}
  \includegraphics[width=\linewidth]{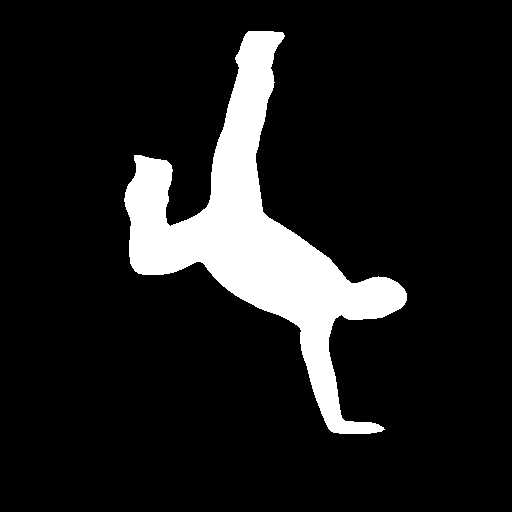}
  \caption{Subject Mask}
  \label{fig:res12}
\end{subfigure}\hfil 
\begin{subfigure}{0.3\textwidth}
  \includegraphics[width=\linewidth]{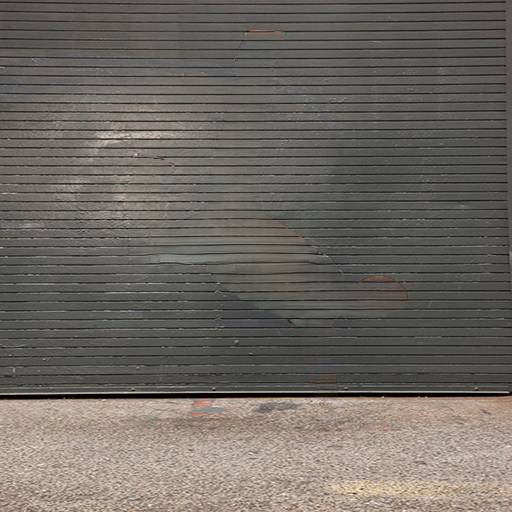}
  \caption{InPainted Image}
  \label{fig:res13}
\end{subfigure}\hfil
\end{figure}

\begin{figure}[H]
    \centering 
\begin{subfigure}{0.3\textwidth}
  \includegraphics[width=\linewidth]{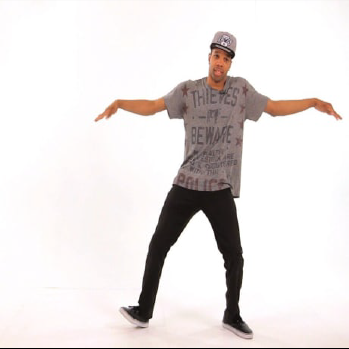}
  \caption{Input Image}
  \label{fig:res11}
\end{subfigure}\hfil 
\begin{subfigure}{0.3\textwidth}
  \includegraphics[width=\linewidth]{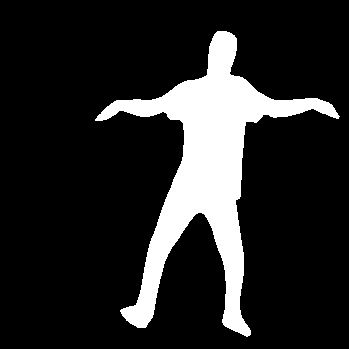}
  \caption{Subject Mask}
  \label{fig:res12}
\end{subfigure}\hfil 
\begin{subfigure}{0.3\textwidth}
  \includegraphics[width=\linewidth]{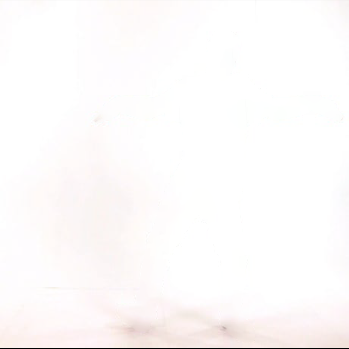}
  \caption{InPainted Image}
  \label{fig:res13}
\end{subfigure}\hfil
\end{figure}

\begin{figure}[H]
    \centering 
\begin{subfigure}{0.3\textwidth}
  \includegraphics[width=\linewidth]{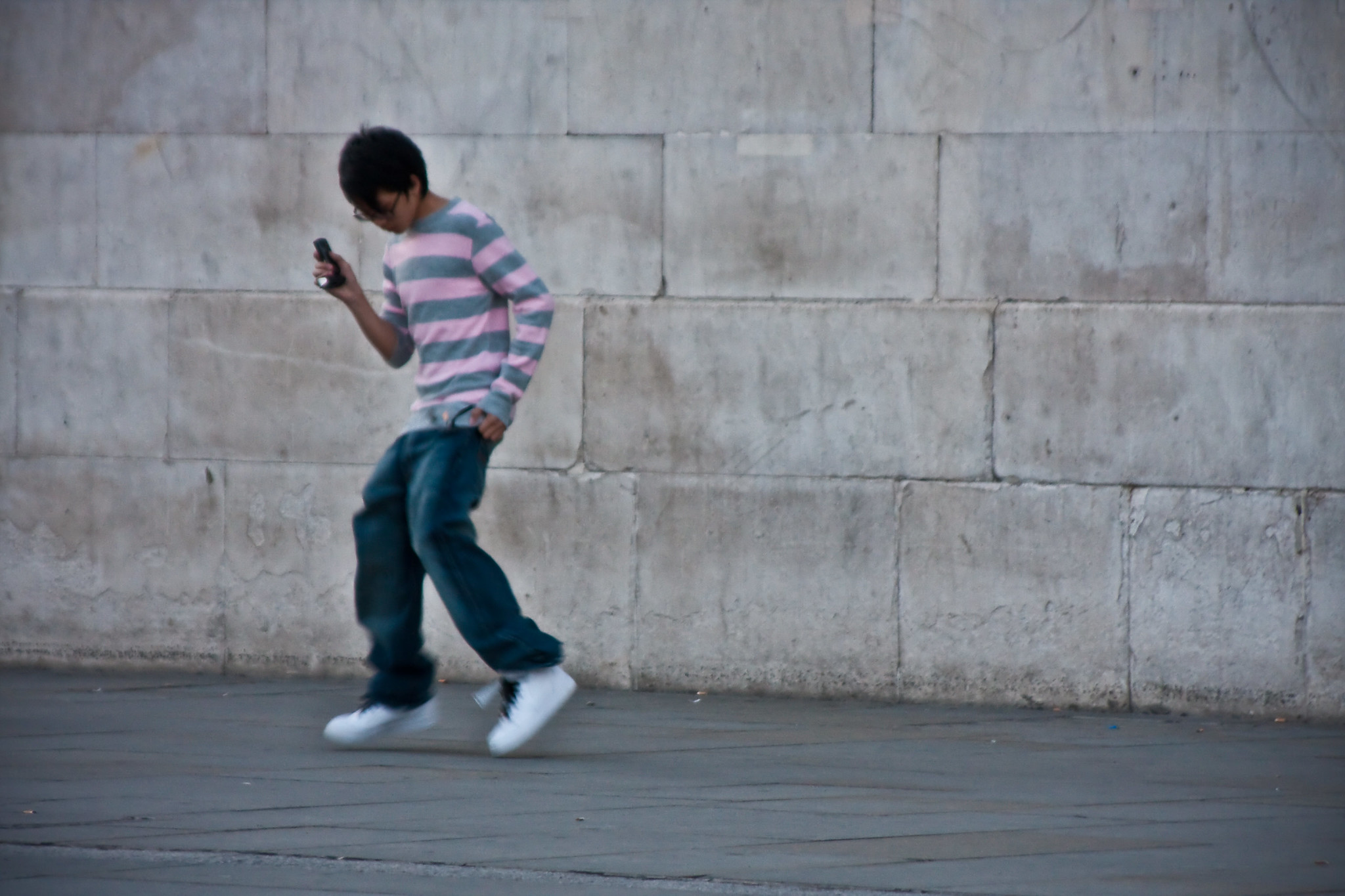}
  \caption{Input Image}
  \label{fig:res11}
\end{subfigure}\hfil 
\begin{subfigure}{0.3\textwidth}
  \includegraphics[width=\linewidth]{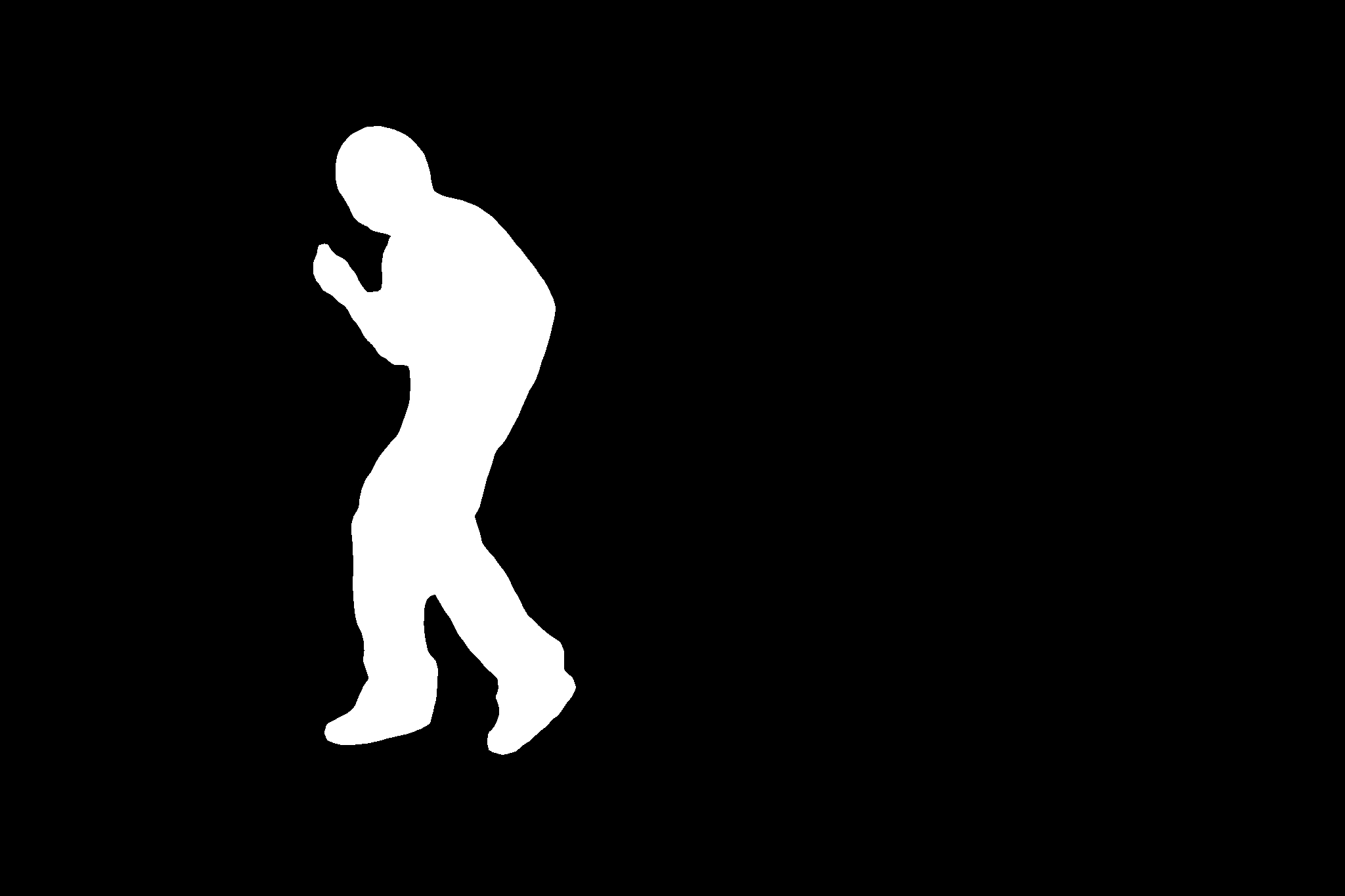}
  \caption{Subject Mask}
  \label{fig:res12}
\end{subfigure}\hfil 
\begin{subfigure}{0.3\textwidth}
  \includegraphics[width=\linewidth]{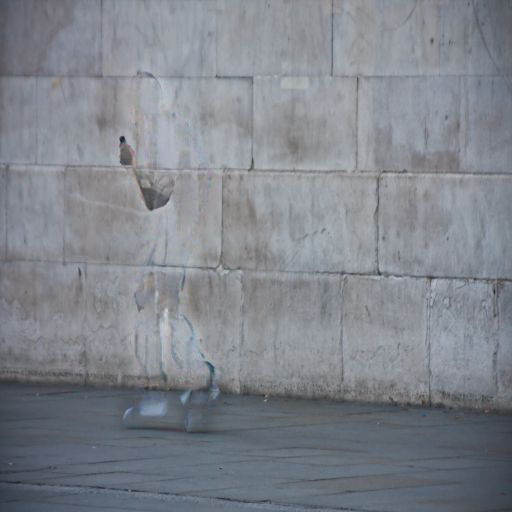}
  \caption{InPainted Image}
  \label{fig:res13}
\end{subfigure}\hfil
\end{figure}

\section{Conclusion}
We present a novel uses case  where we impart motion to a 2D image. We employ  various state-of-the-art solutions to isolated vision problems to give an interactive method of adding dynamics of an otherwise still images. Our method is interactive. One can choose the portion of the image to animate and as well as the poses for animation. The method is also modular, where a better solution to each of the sub-problem can replace the earlier one for improved quality and performance. 
In this work we present a very elementary animation scheme were transformations are done using manipulating vertices of a triangle mesh. In future work we may expand to key-framing and advanced interpolation techniques.

\section{Acknowledgement}
We thank Ankit Phogat and Souymodip Chakraborty who provided insight and expertise that greatly assisted the project. Their willingness to give time so generously has been very much appreciated. Finally, I wish to thank my teachers and parents for their support and encouragement throughout my study.

\bibliographystyle{unsrt}  
\bibliography{references}

\end{document}